\documentclass[sigconf]{aamas}  

\usepackage{booktabs}

\usepackage{url}  
\usepackage{graphicx}  
\usepackage{subcaption}

\usepackage{amsopn}
\usepackage{amsmath}
\usepackage{algorithm}
\usepackage[noend]{algpseudocode}
\usepackage{subcaption}
\usepackage{booktabs}
\usepackage{multirow}

\DeclareMathOperator*{\argmin}{arg\,min}
\newcommand{\mega}{\texttt{MEGA}}
\algrenewcommand\alglinenumber[1]{\tiny #1:}
\newcommand{\footnoteref}[1]{\textsuperscript{\ref{#1}}}

\usepackage{etoolbox}
\makeatletter
\patchcmd{\@verbatim}
  {\verbatim@font}
  {\verbatim@font\small}
  {}{}
\makeatother

\makeatletter
\def\BState{\State\hskip-\ALG@thistlm}
\makeatother

\setcopyright{ifaamas}  
\acmDOI{doi}  
\acmISBN{}  
\acmConference[Technical Report]{}{2018}{ASU}  
\acmYear{2018}  
\copyrightyear{2018}  
\acmPrice{5.00}  

\renewcommand\footnotetextcopyrightpermission[1]{} 

\begin{document}

\title{Balancing Explicability and Explanations}  

\subtitle{Emergent Behaviors in Human-Aware Planning}


\author{Tathagata Chakraborti, Sarath Sreedharan, Subbarao Kambhampati}  
\authornote{The first two authors contributed equally.}
\affiliation{
School of Computing, Informatics, and Decision Systems Engineering\\
Arizona State University, Tempe, AZ 85281 USA\\[1ex]
}
\email{ tchakra2, ssreedh3, rao  @ asu.edu}

\begin{abstract}  
Human aware planning requires an agent to be aware of the intentions,
capabilities and mental model of the human in the loop during its
decision process.  This can involve generating plans that are
explicable to a human observer as well as the ability to
provide explanations  when such plans cannot be
generated. 
In this paper, we bring these two
concepts together and show how an agent can account for both these
needs and achieve a trade-off during the plan generation process
itself by means of a model-space search method \mega. This in effect
provides a comprehensive perspective of what it means for a
decision-making agent to be ``human-aware'' by bringing together
existing principles of planning under the umbrella of a single plan
generation process. We situate our discussion in the context of 
recent work on explicable planning and explanation
generation, and illustrate these concepts in modified versions of two
well-known planning domains, as well as in a demonstration of a robot
involved in a typical search and reconnaissance task with an external
supervisor. Human factor studies in the latter highlight the 
usefulness of the proposed approaches.
\end{abstract}



\maketitle


\section{Introduction}

\begin{figure*}[tbp!]
\centering
\begin{subfigure}[b]{0.59\textwidth}
\centering
\includegraphics[width=0.9\textwidth]{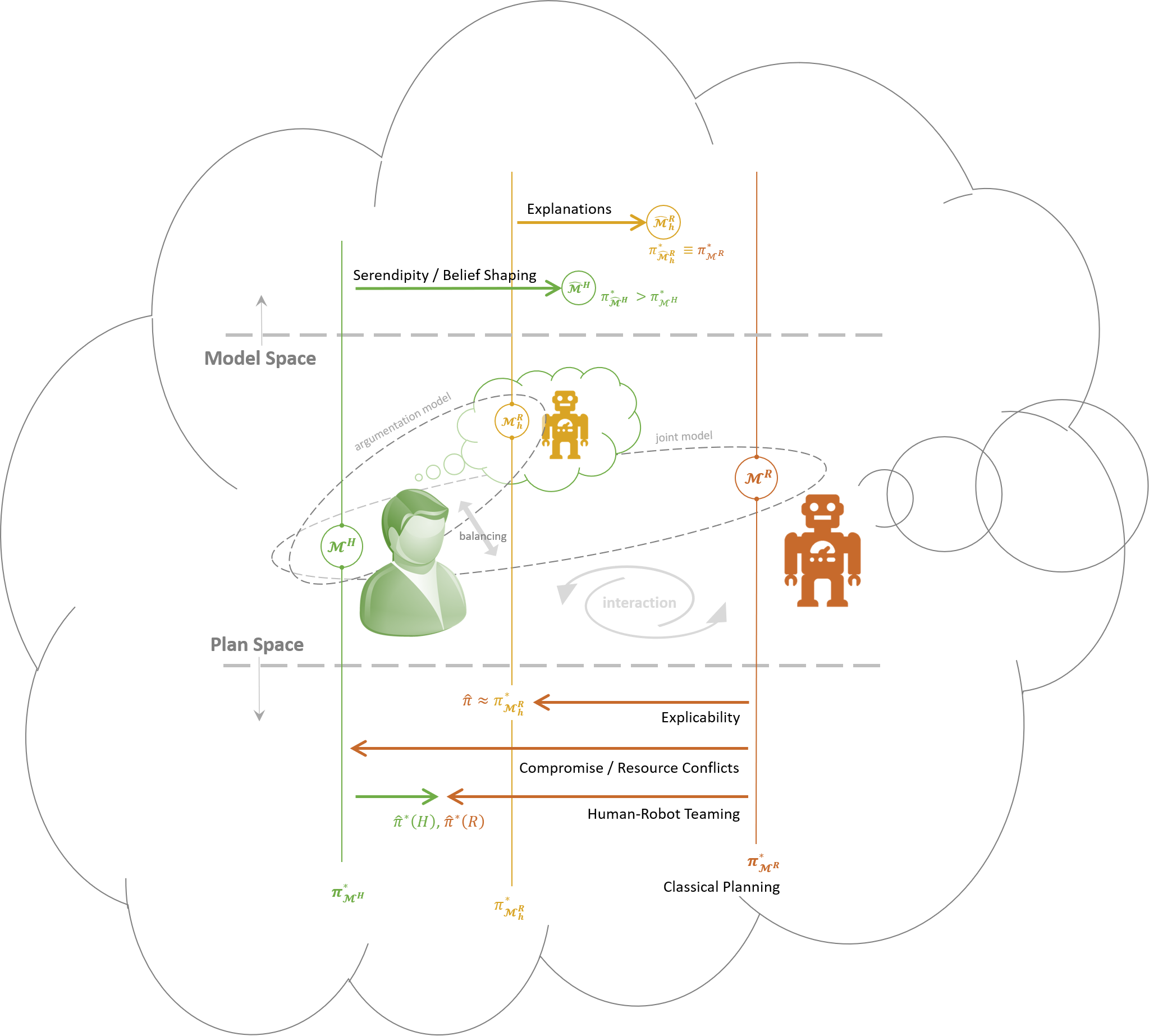}
\vspace{5pt}
\caption{The evolving scope of Human-Aware Planning (HAP)}
\label{hap}
\end{subfigure}
\begin{subfigure}[b]{0.39\textwidth}
\centering
\includegraphics[width=0.9\textwidth]{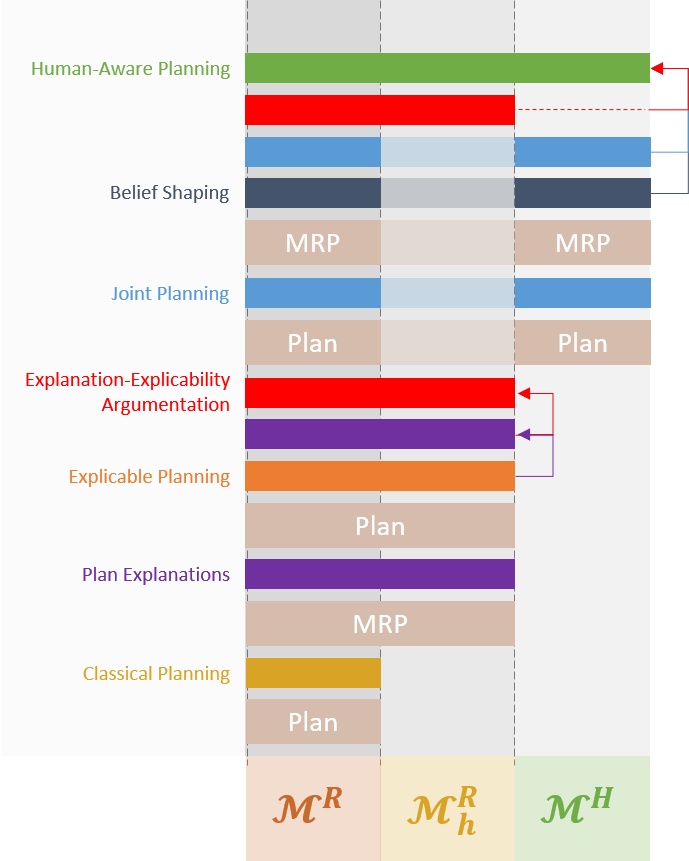}
\vspace{5pt}
\caption{A Subsumption Architecture for HAP}
\label{subsum}
\end{subfigure}
\caption{
The expanding scope of {\em human-aware planning} ({\bf HAP}) acknowledging the need to account for the mental model of the human in the loop in the deliberative process of an autonomous agent. The planner can, for example, choose to bring the human's model closer to the ground truth using explanations via a process called {\em model reconciliation} (MRP) so that an otherwise inexplicable plan makes sense in the human's updated model or it can compute explicable plans which are closer to the human's expectation. These capabilities can be stacked to realize more and more complex behavior -- in this paper we will concentrate on the explicability versus explanation trade-off as a form of argumentation during human-aware planning.
}
\label{fig:fig}
\end{figure*}

It is often useful for a planning agent while {\em interacting} with a human in the loop to use, in the process of its deliberation, not only the model $\mathcal{M}^R$ of the task it has on its own, but also the model $\mathcal{M}^R_h$ that the human thinks it has (refer to Figure \ref{hap}).
This mental model of the human \cite{chakraborti2017ai} is in addition to the physical model of the human.
This is, in essence, the fundamental thesis of the recent works on plan explanations \cite{explain} and explicable planning \cite{exp-yu}, summarized under the umbrella of {\em multi-model planning}, 
and is in addition to the originally studied {\em human-aware planning} (HAP) problems where actions of the human (and hence the actual {\em human model} and the robot's belief of it) are also involved in the planning process. 
The need for explicable planning or plan explanations in fact occur when these two models -- $\mathcal{M}^R$ and $\mathcal{M}^R_h$ -- diverge. 
This means that the optimal plans in the respective models -- $\pi^*_{\mathcal{M}^R}$ and $\pi^*_{\mathcal{M}^R_h}$ -- may not be the same and hence optimal behavior of the robot in its own model is inexplicable to the human in the loop. 
In the explicable planning process, the robot produces a plan $\widehat{\pi}$ that is closer to the human's expected plan, i.e. $\widehat{\pi} \approx \pi^*_{\mathcal{M}^R_h}$. 
In the explanation process, the robot instead attempts to update the human's mental model to an intermediate model $\widehat{\mathcal{M}}^R_h$ in which the robot's original plan is {\em equivalent} (with respect to a metric such as cost or similarity) to the optimal and hence explicable, i.e. $\pi^*_{\mathcal{M}^R} \equiv \pi^*_{\widehat{\mathcal{M}}^R_h}$. 


Until now, these two processes of plan explanations and explicability have remained separate in so far as their role in an agent's deliberative process is considered - i.e. a planner either generates an explicable plan to the best of its ability or it produces explanations of its plans where they required. However, there may be situations where a combination of both provide a much better course of action -- if the expected human plan is too costly in the planner's model (e.g. the human might not be aware of some safety constraints) or the cost of communication overhead for explanations is too high (e.g. limited communication bandwidth). Consider, for example, a human working with a robot that has just received a software update allowing it to perform new complex maneuvers. Instead of directly trying to conceive all sorts of new interactions right away that might end up spooking the user, the robot could instead reveal only certain parts of the new model while still using its older model (even though suboptimal) for the rest of the interactions so as to slowly reconcile the drifted model of the user. This is the focus of the current paper where we try to attain the sweet spot between plan explanations and explicability.


\subsection{Related Work}

As AI agents become pervasive in our daily lives, the need for such agents to be cognizant of the beliefs and expectations of the humans in their environment has been well documented \cite{hilp}. From the perspective of task planning, depending on the extent of involvement of the human in the life cycle of a plan, 
work in this direction has ranged on a spectrum of ``human-aware planning'' \cite{alami2006toward,alami2014human,tist,grandpa,tomic,cirillo2010planning,iros,aamas} where a robot passively tries to account for the plans of humans cohabiting its workspace, to ``explicable planning'' \cite{zhang2016plan,exp-yu,exp-anagha,Dragan–2013–7671} where a robot generates plans that are explicable or predictable to a human observer, to ``plan explanations'' \cite{explain,langley2017explainable,danmaga} where the agent uses explanations to bring the human (who may have a different understanding of the agent's abilities) on to the same page, to ``human-in-the-loop planning'' \cite{allen1994mixed,ferguson1996trains,ai2004mapgen,manikondaherding,radar} in general where humans and planners are participating in the plan generation and/or execution process together. 

\subsubsection{The Evolving Scope of Human-Aware Planning}

The ongoing efforts to make planning more ``human-aware'' is illustrated in Figure \ref{hap} -- initial work on this topic had largely focused on incorporating an agent's understanding of the human model $\mathcal{M}^H$ into its decision making process.
Since then the importance of considering the human's understanding $\mathcal{M}^R_h$ of the agent's actual model $\mathcal{M}^R$ in the planning process has also been acknowledged, sometimes implicitly \cite{alami2014human} and later explicitly \cite{exp-yu,explain}. 
These considerations engender interesting behaviors both in the space of plans and models. 
For example, in the model space, the modifications to the human mental model $\mathcal{M}^R_h$ is used for explanations in \cite{explain} while reasoning over the actual model $\mathcal{M}^H$ can reveal interesting behavior by affecting the belief state of the human, such as in planning for serendipity \cite{iros}. In the plan space, a human-aware agent can use $\mathcal{M}^H$ and $\mathcal{M}^R_h$ to compute joint plans for teamwork \cite{talamadupula2014coordination} or generate behavior that conforms to the human's preferences \cite{grandpa,aamas} and expectations \cite{zhang2016plan,exp-yu,exp-anagha}. 
From the point of view of the planner, this is, in a sense, an asymmetric epistemic setting with single-level nested beliefs over its models. 
Indeed, existing literature on epistemic reasoning \cite{hanheide2015robot,muise2016planning,miller2017logics} can also provide interesting insights in the planning process of an agent in these settings.

\subsubsection{A Subsumption Architecture for HAP}

These different forms of behavior can be {\em composed} to form more and more sophisticated forms of human-aware behavior. This hierarchical composition of behaviors can be viewed in the form of a subsumption architecture for human-aware planning, similar in motivation to \cite{brooks1986robust}. This is illustrated in Figure~\ref{subsum}. The basic reasoning engines are the Plan and MRP (Model Reconciliation) modules. The former accepts model(s) of planning problems and produces a plan, the latter accepts the same and an produces a new model. The former operates in plan space and gives rise to classical, joint and explicable planning depending on the models it is operating on, while the latter operates in model space to produce explanations and belief shaping behavior. These are then composed to form argumentation modules for trading of explanations and explicability (which is the topic of the current paper) and human-aware planning in general.


\subsubsection{The Explicability-Explanation Trade-off}

From the perspective of design of autonomy, this trade-off has two important implications -- (1) the agent can now not only explain but also {\em plan} in the multi-model setting with the trade-off between compromise on its optimality and possible explanations in mind; and (2) the argumentation process is known to be a crucial function of the reasoning capabilities of humans \cite{argue}, and now by extension of autonomous agents as well, as a result of algorithms we develop here to incorporate the explanation generation process into the agent's decision making process itself.
General argumentation frameworks for resolving disputes over plans have indeed been explored before \cite{belesiotis2010agreeing,emele2011argumentation}. 
Our work can be seen as the specific case where the argumentation process is over a set of constraints that prove the correctness and quality of plans by considering the cost of the argument specifically as it relate to the trade-off in plan quality and the cost of explaining that plan. This is the first of its kind algorithm that can achieve this in the scope of plan explanations and explicable in presence of model differences with the human.






\section*{Human-Aware Planning Revisited}

The problem formulation closely follows that introduced in \cite{explain}, reproduced here for clarity of methods built on the same definitions. 

\subsubsection*{A Classical Planning Problem} is a tuple $\mathcal{M} = \langle \mathcal{D}, \mathcal{I}, \mathcal{G} \rangle$\footnote{Note that the ``model of a planning problem'' includes the action model {\em as well as the initial and goal states of an agent.}} 
with domain $\mathcal{D} = \langle F, A\rangle$ - where $F$ is a set of fluents that define a state $s \subseteq F$, and $A$ is a set of actions - and initial and goal states $\mathcal{I}, \mathcal{G} \subseteq F$. 
Action $a \in A$ is a tuple $\langle c_a, \textit{pre}(a), \textit{eff}^\pm(a)\rangle$ where $c_a$ is the cost, and $\textit{pre}(a), \textit{eff}^\pm(a) \subseteq F$ are the preconditions and add/delete effects, i.e. $\delta_{\mathcal{M}}(s, a) \models \bot \textit{ if } s\not\models \textit{pre}(a); \textit{ else } \delta_{\mathcal{M}}(s, a) \models s \cup \textit{eff}^+(a) \setminus \textit{eff}^-(a)$ where $\delta_{\mathcal{M}}(\cdot)$ is the transition function. 
The cumulative transition function is 
$\delta_{\mathcal{M}}(s,\langle a_1, a_2, \ldots, a_n \rangle) = \delta_{\mathcal{M}}(\delta_{\mathcal{M}}(s, a_1),\langle a_2, \ldots, a_n \rangle)$.

\vspace{5pt}
\noindent The solution to the planning problem is a sequence of actions or a (satisficing) {\em plan} $\pi = \langle a_1, a_2, \ldots, a_n \rangle$ such that $\delta_{\mathcal{M}}(\mathcal{I}, \pi) \models \mathcal{G}$. 
The cost of a plan $\pi$ is $C(\pi, \mathcal{M}) = \sum_{a\in\pi}c_a$ if $\delta_{\mathcal{M}}(\mathcal{I}, \pi) \models \mathcal{G}$; $\infty$ otherwise. 
The cheapest plan $\pi^* = \argmin_{\pi} C(\pi,\mathcal{M})$ is the (cost) optimal plan. We refer to this cost as $\mathcal{M}$ as $C_\mathcal{M}^*$.

\subsubsection*{A Human-Aware Planning (HAP) Problem} is given by the tuple $\Psi = \langle \mathcal{M}^R, \mathcal{M}^H, \mathcal{M}^R_h \rangle$ where $\mathcal{M}^R = \langle D^R, \mathcal{I}^R, \mathcal{G}^R \rangle$ is the planner's model of a planning problem, while $\mathcal{M}^H = \langle D^H, \mathcal{I}^H, \mathcal{G}^H \rangle$ and $\mathcal{M}^R_h = \langle D^R_h, \mathcal{I}^R_h, \mathcal{G}^R_h \rangle$ are respectively the planner's estimate of the human's model and the human's understanding of its own model. 

\vspace{5pt}
\noindent The solution to the human-aware planning problem is a joint plan \cite{iros} $\pi = \langle a_1, a_2, \ldots, a_n \rangle; ~a_i \in \{D^R \cup D^H_r\}$ such that $\delta_{\Psi}(\mathcal{I}^R \cup \mathcal{I}^H_r, \pi) \models \mathcal{G}^R \cup \mathcal{G}^H_r$. 
The robot's component in the plan is referred to as $\pi(R) = \langle a_i~|~a_i \in \pi \wedge D^R \rangle$. For the purposes of this paper, we ignore the robot's belief of the human model, i.e. $\mathcal{M}^H_r = \langle \{\}, \{\}, \{\} \rangle$ -- in effect, making the human an observer only or a passive consumer of the plan -- and focus instead on the challenges involves in planning with the human's model of the planner. 
Planning with the human model has indeed been studied extensively in the literature, as noted above, and this assumption does not change in any way the relevance of the work here. Specifically, the following concepts are built on top of the joint planning problem -- e.g. an explicable plan in this paper would, in the general sense, correspond to the robot's component in the joint plan being explicable to the human in the loop. 
Thus, for the purposes of this paper, we have $\pi(R) \equiv \pi$; 
without loss of generality, we focus on the simplified 
setting with only the model of the planner and the human's approximation of it.

\subsection*{Explicable Planning}
In ``explicable planning" a solution to the human-aware planning problem is a plan $\pi$ such that (1) it is executable (but may no longer be optimal) in the robot's model but is (2) ``closer'' to the expected plan in the human's model, {\em given a particular planning problem} -- 

\begin{itemize}
\item[(1)] $\delta_{\mathcal{M}^R}(\mathcal{I}^R, \pi) \models \mathcal{G}^R$; and
\item[(2)] $C(\pi, \mathcal{M}^R_h) \approx C^*_{\mathcal{M}^R_h}$.
\end{itemize}

\noindent ``Closeness'' or distance to the expected plan is modeled here in terms of cost optimality, but in general this can be any preference metric like plan similarity.
In existing literature \cite{exp-yu,zhang2016plan,exp-anagha} this has been usually achieved by modifying the search process so that the heuristic that guides the search is driven by the robot's knowledge of the human's mental model. 
Such a heuristic can be either derived directly \cite{exp-anagha} from the human's model (if it is known)  or learned \cite{exp-yu} through interactions in the form of affinity functions between plans and their purported goals.

\subsection*{Plan Explanations}

The other approach would be to (1) compute optimal plans in the planner's model as usual, but also provide an explanation (2) in the form of a model update to the human so that (3) the same plan is now also optimal in the human's updated model of the problem. 
Thus, a solution involves a plan $\pi$ and an explanation $\mathcal{E}$ such that --

\begin{itemize}
\item[(1)] $C(\pi, \mathcal{M}^R) = C^*_{\mathcal{M}^R}$;
\item[(2)] $\mathcal{\widehat{M}}^R_h \longleftarrow \mathcal{M}^R_h + \mathcal{E}$; and
\item[(3)] $C(\pi, \mathcal{\widehat{M}}^R_h) = C^*_{\mathcal{\widehat{M}}^R_h}$. 
\end{itemize}

\noindent Note that here a model update, as indicated by the $+$ operator 
may include a correction to the belief (goals or state information) as well as information pertaining to the action model itself.
In \cite{explain} the authors explored various ways of generating such solutions -- including methods to minimize the lengths of the explanations given as a result. However, this was done in an after-the-fact fashion, i.e. the optimal plan was already generated and it was just a matter of finding the best explanation for it. This not only ignores the possibility of finding better plans (that are equally optimal) with smaller explanations, but also avenues of compromise in a manner we discussed previously whereby the planner sacrifices its optimality to further reduce overhead in the explanation process. 

\section*{\mega}

We bring the notions of explicability and explanations together in a novel planning technique \mega~(Multi-model Explanation Generation Algorithm) that trades off the relative cost of explicability to providing explanations during the plan generation process itself\footnote{As in \cite{explain} we assume that the human mental model is known and has the same computation power (\cite{explain} also suggests possible ways to address these issues, the same discussions apply here as well). Also refer to the discussion on model learning later.
}. 
The output of \mega~is a plan $\pi$ and an explanation $\mathcal{E}$ such that (1) $\pi$ is executable in the robot's model, and with the explanation (2) in the form of model updates it is (3) optimal in the updated human model while (4) the cost (length) of the explanations, and the cost of deviation from optimality in its own model to be explicable to the human, is traded off according to a constant $\alpha$ -- 
\begin{itemize}
\item[(1)] $\delta_{\mathcal{M}^R}(\mathcal{I}^R, \pi) \models \mathcal{G}^R$;
\item[(2)] $\mathcal{\widehat{M}}^R_h \longleftarrow \mathcal{M}^R_h + \mathcal{E}$;
\item[(3)] $C(\pi, \mathcal{\widehat{M}}^R_h) = C^*_{\mathcal{\widehat{M}}^R_h}$; and
\item[(4)] $\pi = \argmin_{\pi}~\{~|\mathcal{E}|~+~\alpha \times |~C(\pi, \mathcal{M}^R) -  C^*_{\mathcal{M}^R}~|~\}$.
\end{itemize}

\noindent Clearly, with higher values of $\alpha$ the planner will produce plans that require more explanation, with lower $\alpha$ it will generate more explicable plans. Thus, with the help of this hyperparameter $\alpha$, an autonomous agent can deliberate over the trade-off in the costs it incurs in being explicable to the human (second minimizing term in (4)) versus explaining its decisions (first minimizing term in (4)). 
Note that this trade-off is irrespective of the cognitive burden of those decisions on the human in the loop. 
For example, for a robot in a collapsed building during a search and rescue task, or the rover on Mars, may have limited bandwidth for communication and hence prefer to be explicable instead instead.

We employ a {\em model space} $A^*$ search (Algorithm \ref{algo}) to compute the expected plan and explanations for a given value of $\alpha$. 
Similar to \cite{explain} we define a state representation over planning problems with a mapping function $\Gamma :~^a\mathcal{M} \mapsto \mathcal{F}$ which represents a planning problem by transforming every condition in it into a predicate. 
The set $\Lambda$ of actions contains unit model change actions $\lambda : \mathcal{F} \rightarrow \mathcal{F}$ which make a single change to a domain at a time.

We start by initializing the min node tuple ($\mathcal{N}$) with the human mental model and an empty explanation. For each new possible model we come across during our model space search, we test if the objective value of the new node is smaller than the current min node. 
We stop the search once we identify a model that is capable of producing a plan that is also optimal in the robot's own model. This is different from the stopping condition used by the original MCE-search\footnote{\label{mce}An MCE or a {\em minimally complete explanation} is the shortest model update so that {\em a given plan} optimal in the robot model is also optimal in the updated human model.} in \cite{explain}, where the authors are just trying to identify the first node where the given plan is optimal. 

\subsubsection*{Property 1} \mega~yields the smallest possible explanation for a given human-aware planning problem.

\vspace{5pt}
\noindent This means that with a high enough $\alpha$ (see below) the algorithm is guaranteed to compute the best possible plan for the planner as well as the smallest explanation associated with it. 
This is by construction of the search process itself, i.e. the search only terminates after the all the nodes that allow $C(\pi, \mathcal{\widehat{M}}^R_h) = C^*_{\mathcal{\widehat{M}}^R_h}$ have been exhausted. 
This is beyond what is offered by the model reconciliation search in \cite{explain}, which only computes the smallest explanation {\em given} as a plan that is optimal in the planner's model.

\subsubsection*{Property 2} $\alpha = |~\mathcal{M}^R~\Delta~\mathcal{M}^R_h~|$ yields the most optimal plan in the planner's model along with the minimal explanation possible given a human-aware planning problem.

\vspace{5pt}
\noindent This is easy to see, since with $\forall \mathcal{E},~|\mathcal{E}| \leq |~\mathcal{M}^R~\Delta~\mathcal{M}^R_h~|$, the latter being the total model difference, the penalty for departure from explicable plans is high enough that the planner must choose from possible explanations only (note that the explicability penalty is always positive until the search hits the nodes with $C(\pi, \mathcal{\widehat{M}}^R_h) = C^*_{\mathcal{\widehat{M}}^R_h}$, at which point onwards the penalty is exactly zero). In general this works for any $\alpha \geq |MCE|$ but since an MCE will only be known retrospectively after the search is complete, the above condition suffices since the entire model difference is known up front and is the largest possible explanation in the worst case.

\begin{algorithm}[!tp]
\scriptsize
\caption{\mega}
\label{algo}
\begin{algorithmic}[1]
\Procedure{\mega-Search}{}
\vspace{2pt} 
\BState \emph{Input}:~~~~HAP $\Psi = \langle \mathcal{M}^R, \mathcal{M}^R_h \rangle$, $\alpha$
\BState \emph{Output}: Plan $\pi$ and Explanation $\mathcal{E}$
\vspace{2pt} 
\BState \emph{Procedure}:  
\vspace{2pt} 
\State fringe~~~~~~~~$\leftarrow$ \texttt{Priority\_Queue()}
\State c\_list~~~~~~~~~~$\leftarrow$ \{\}  \Comment{\textcolor{black}{Closed list}}
\State $\mathcal{N}_{min}$~~~~~~$\leftarrow\langle\mathcal{M}^R_h,\{\}\rangle$ \Comment{\textcolor{black}{Node with minimum objective value}}
\State $\pi^*_R$~~~~~~~~~~~~$\leftarrow \pi^*$ \Comment{\textcolor{black}{Optimal plan being explained}}
\State $\pi^R_h$~~~~~~~~~~~~$\leftarrow \pi$ s.t. $C(\pi, \mathcal{M}^R_h) = C^{*}_{\mathcal{M}^R_h}$ \Comment{\textcolor{black}{Plan expected by human}}
\vspace{2pt} 
\State $\text{fringe.push}(\langle \mathcal{M}^R_h, \{\}\rangle,~\text{priority} = 0)$
\vspace{2pt} 
\While{True}
\vspace{2pt} 
\State $\langle \widehat{\mathcal{M}}, \mathcal{E} \rangle, c \leftarrow \text{fringe.pop}(\widehat{\mathcal{M}})$
\If{$\textrm{OBJ\_VAL}(\langle \widehat{\mathcal{M}}, \mathcal{E} \rangle) \leq \textrm{OBJ}\_\textrm{VAL}(\mathcal{N}_{min}) $}
\State $\mathcal{N}_{min}$~~~~~$\leftarrow\langle \widehat{\mathcal{M}}, \mathcal{E} \rangle$ \Comment{\textcolor{black}{Update min node}}
\EndIf
\If{$C(\pi^*_{\widehat{\mathcal{M}}}, \mathcal{M}^{R}) = C^{*}_{\mathcal{M}^{R}}$}
\State $\langle\mathcal{M}_{min}, \mathcal{E}_{min}\rangle~\leftarrow~\mathcal{N}_{min}$
\State \Return $\langle\pi_{\mathcal{M}_{min}},\mathcal{E}_{min}\rangle$  \Comment{\textcolor{black}{If $\pi^*_{\widehat{\mathcal{M}}}$ is optimal in $\mathcal{M}^R$}}
\Else
\State c\_list $\leftarrow$ c\_list $\cup~\widehat{\mathcal{M}}$
\vspace{2pt} 
\For{$f \in \Gamma(\widehat{\mathcal{M}})~\setminus~\Gamma(\mathcal{M}^R)$} \Comment{\textcolor{black}{Models that satisfy condition 1}}
\State $\lambda \leftarrow \langle 1, \{\widehat{\mathcal{M}}\}, \{\}, \{f\} \rangle$ \Comment{\textcolor{black}{Removes f from $\widehat{\mathcal{M}}$}}
\If{$\delta_{\mathcal{M}^R_h,\mathcal{M}^R}(\Gamma(\widehat{\mathcal{M}}), \lambda) \not\in \text{c\_list}$}
\State $\text{fringe.push}(\langle \delta_{\mathcal{M}^R_h,\mathcal{M}^R}(\Gamma(\widehat{\mathcal{M}}), \lambda),~\mathcal{E}~\cup~\lambda \rangle,~c + 1)$
\EndIf
\EndFor
\vspace{2pt} 
\For{$f \in \Gamma(\mathcal{M}^R)~\setminus~\Gamma(\widehat{\mathcal{M}})$} \Comment{\textcolor{black}{Models that satisfy condition 2}}
\State $\lambda \leftarrow \langle 1, \{\widehat{\mathcal{M}}\}, \{f\}, \{\} \rangle$ \Comment{\textcolor{black}{Adds f to $\widehat{\mathcal{M}}$}}
\If{$\delta_{\mathcal{M}^R_h,\mathcal{M}^R}(\Gamma(\widehat{\mathcal{M}}), \lambda) \not\in \text{c\_list}$}
\State $\text{fringe.push}(\langle \delta_{\mathcal{M}^R_h,\mathcal{M}^R}(\Gamma(\widehat{\mathcal{M}}), \lambda),~\mathcal{E}~\cup~\lambda \rangle,~c + 1)$
\EndIf
\EndFor
\EndIf
\EndWhile
\Procedure{OBJ\_VAL}{$\langle \widehat{\mathcal{M}}, \mathcal{E} \rangle$}
\State \Return~~$|\mathcal{E}|~+~\alpha \times |~C(\pi_{\mathcal{\widehat{M}}}^*,\mathcal{M}^R) -  C^*_{\mathcal{M}^R}~|$
\EndProcedure
\EndProcedure
\end{algorithmic}
\end{algorithm}

\subsubsection*{Property 3} $\alpha = 0$ yields the most explicable plan.

\vspace{5pt}
\noindent Under this condition, the planner has to minimize the cost of explanations only. Of course, at this point it will produce the plan that requires the shortest explanation, and hence the most explicable plan. Note that this is distinct from just computing the optimal plan in the human's model, since such a plan may not be executable in the planner's model so that some explanations are required even in the worst case. This is also a welcome additions to the explicability only view of plan generation introduced in \cite{exp-yu,exp-anagha,zhang2016plan}, where the human model only also guides the plan generation process instead of doing so directly, though none of these works provided any insight into how to make the remainder of the model reconciliation possible in such cases, as done here with the explanations associated with the generated plans. 

\subsubsection*{Property 4} \mega-search is required only once per problem, and is independent of $\alpha$.

\vspace{5pt}
\noindent Algorithm \ref{algo} terminates only after all the nodes containing a minimally complete explanation\footnoteref{mce} have been explored. This means that for different values of $\alpha$, the agent only needs to post-process the nodes with the new objective function in mind. Thus, a large part of the reasoning process for a particular problem can be pre-computed.

\section{Evaluations}
We will now provide internal evaluations of \mega~in modified versions of two well-known IPC domains \texttt{Rover} and \texttt{Barman} \cite{ipc} demonstrating the trade-off in the cost and computation time of plans with respect to varying size of the model difference and the hyper-parameter $\alpha$ and follow it up with a demonstration of \mega~ in action on a robot in a search and reconnaissance domain.  
Finally, we will report on human factor studies on how this trade-off is received by users.
The code and the domain models will be available after the double-blind review process is over.

\begin{table}
\tiny
\centering
  \begin{tabular}{r|c|c|c|c|c|c|c}
    \toprule
    \multirow{2}{*}{Domain Name} &
    \multirow{2}{*}{Problem} &
     \multicolumn{2}{c|}{$\Delta = 2$} &
     \multicolumn{2}{c|}{$\Delta = 7$} &
      \multicolumn{2}{c}{$\Delta = 10$}\\
      & 
	      & {$|\mathcal{E}|$} & {Time (secs)}& {$|\mathcal{E}|$} & {Time (secs)} & {$|\mathcal{E}|$} & Time (secs) \\
      \midrule
      \midrule
    \multirow{3}{*}{\texttt{Rover}} &p1& 0 & 1.22 & 1 & 5.83 & 3   &  143.84   \\ 
     & p2 & 1&1.79 & 5 &125.64&6  & 1061.82 \\
        & p3 &0 & 8.35 &2&10.46& 3  & 53.22  \\
      \midrule
       \multirow{3}{*}{\texttt{Barman}} & p1& 2 & 18.70  &6&163.94&  6 & 5576.06 \\
       &p2&2 &2.43&  4&57.83&6 & 953.47\\
       &p3&2 &45.32&  5&4183.55&6 & 5061.50\\
    \bottomrule
    \bottomrule
  \end{tabular}
\vspace{10pt}
\caption{
Computation time for human-aware plans in \texttt{Rover} and \texttt{Barman} domains along with the length of explanations.
}
\label{times}
\vspace{-10pt}
\end{table}

\subsection{Empirical Results: Cost Trade-off}

The value of $\alpha$ determines how much an agent is willing to sacrifice its own optimality versus the cost of explaining a (perceived) suboptimal plan to the human. 
In the following, we illustrate this trade-off on modified versions of two well-known IPC domains.

\subsubsection{The Rover (Meets a Martian) Domain}

Here the Mars Rover has a model as described in the IPC domain, but has gone an update whereby it can carry all the rock and soil samples needed for a mission at the same time. This means that it does not need to empty the store before collecting new rock and soil samples anymore so that the new action definitions for $\texttt{sample\_soil}$ and $\texttt{sample\_rock}$ no longer contain the precondition $\texttt{(empty ?s)}$. 

During its mission it runs across a Martian who is unaware of the robot's expanded storage capacity, and has an older, extremely cautious, model of the rover it has learned while spying on it from its cave. 
It believes that any time we collect a rock sample, we also need to collect a soil sample and need to communicate this information to the lander. 
It also believes that before the rover can perform \texttt{take\_image} action, it needs to send the soil data and rock data of the waypoint from where it is taking the image. 
Clearly, if the rover was to follow this model in order not to spook the Martians, 
it will end up spending a lot of time performing unnecessary actions (like dropping old samples and collecting unnecessary samples).
For example, if the rover is to communicate an image of an objective \texttt{objective2}, all it needs to do is move to a waypoint (\texttt{waypoint3}) from where \texttt{objective2} is visible and perform the action --

\vspace{5pt}
{
\begin{verbatim}
(take_image waypoint3 objective2 camera0 high_res)
\end{verbatim}
}

\begin{figure}[tbp!]
\centering
\begin{subfigure}[b]{0.9\columnwidth}
\includegraphics[width=\columnwidth]{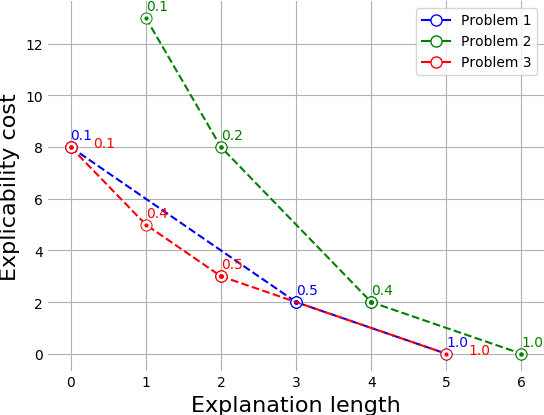}
\caption{The Rover (Meets a Martian) Domain}
\label{martian}
\vspace{5pt}
\end{subfigure}
\begin{subfigure}[b]{0.9\columnwidth}
\includegraphics[width=\columnwidth]{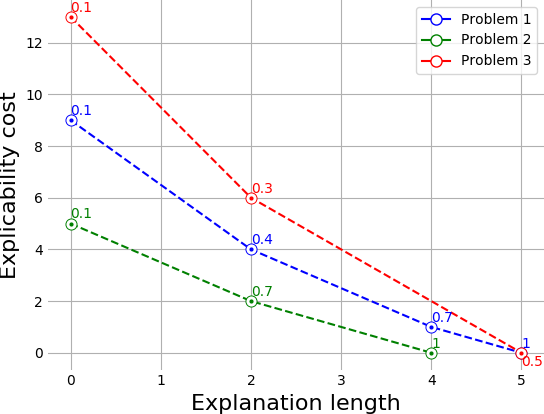}
\caption{The Barman (in a Bar) Domain}
\label{barman}
\end{subfigure}
\caption{Trade-off between explicability versus explanation cost for plans produced at different values of $\alpha$.}
\label{graph-fig}
\end{figure}

\noindent If the rover was to produce a plan that better represents the Martian's expectations, it would look like --

\vspace{5pt}
{
\begin{verbatim}
(sample_soil store waypoint3)
(communicate_soil_data general waypoint3 waypoint3 waypoint0)
(drop_off store)
(sample_rock store waypoint3)
(communicate_rock_data general waypoint3 waypoint3 waypoint0)
(take_image waypoint3 objective1 camera0 high_res)
\end{verbatim}
}

Now if the rover chose to directly use an MCE it could end up explaining up to six different model differences based on the problem and the plan under execution. In some case, this may be acceptable, but in others, it may make more sense for the rover to bear the extra cost rather than laboriously walking through all the updates with an impatient Martian. \mega~lets us naturally model these scenarios through the use of the $\alpha$ parameter -- the rover would choose to execute the Martian's expected optimal plan when the $\alpha$ parameter is set to zero (which means the rover does not care about the extra cost it needs to incur to ensure that the plan makes sense to the Martian with the least explaining involved). 

Figure \ref{graph-fig} shows how the explicability cost and explanation cost varies for three typical problem instances in this domain. The algorithm starts converging to the smallest possible MCE, when $\alpha$ is set to one. For smaller $\alpha$, \mega~ chooses to save explanation costs by choosing more expensive (and explicable) plans.

\subsubsection{The Barman (in a Bar) Domain}

Here, the brand new two-handed Barman robot is wowing onlookers with its single-handed skills, even as its admirers who may be unsure of its capabilities expect, much like in the original IPC domain, that it is required to have one hand free to perform actions like \texttt{fill-shot}, \texttt{refill-shot}, \texttt{shake} etc. 
This means that to make a single shot of a cocktail with two shots of the same ingredient with three shots and one shaker, the human expects the robot to execute the following plan --

\vspace{5pt}
{
\begin{verbatim}
(fill-shot shot2 ingredient2 left right dispenser2)
(pour-shot-to-used-shaker shot2 ingredient3 shaker1 left l1 l2)
(refill-shot shot2 ingredient3 left right dispenser3)
(pour-shot-to-used-shaker shot2 ingredient3 shaker1  left l1 l2)
(leave left shot2)
(grasp left shaker1)
\end{verbatim}
}

The robot can, however, directly start by picking both the shot and the shaker and does not need to put either of them down while making the cocktail.
Similar to the Rover domain, we again illustrate on three typical problems from the barman domain (Figure \ref{graph-fig}) how at lower values of $\alpha$ the robot choose to perform plans that require less explanation. As $\alpha$ increases the algorithm produces plans that require larger explanations with the explanations finally converging at the smallest MCE required for that problem.


\subsection{Empirical Results: Computation Time}

Contrary to classical notions of planning that occurs in state or plan space, we are now planning in the model space, i.e. every node in the search tree is a new planning problem. 
As seen in Table \ref{times} this becomes quite time consuming with increasing number of model differences between the human and the robot, even as there are significant gains to be had in terms of minimality of explanations, and the reduction in cost of explicable plans as a result of it. 
This motivates the need for developing approximations and heuristics \cite{explain} for the search for multi-model explanations.

\begin{figure}[tbp!]
\includegraphics[width=\columnwidth]{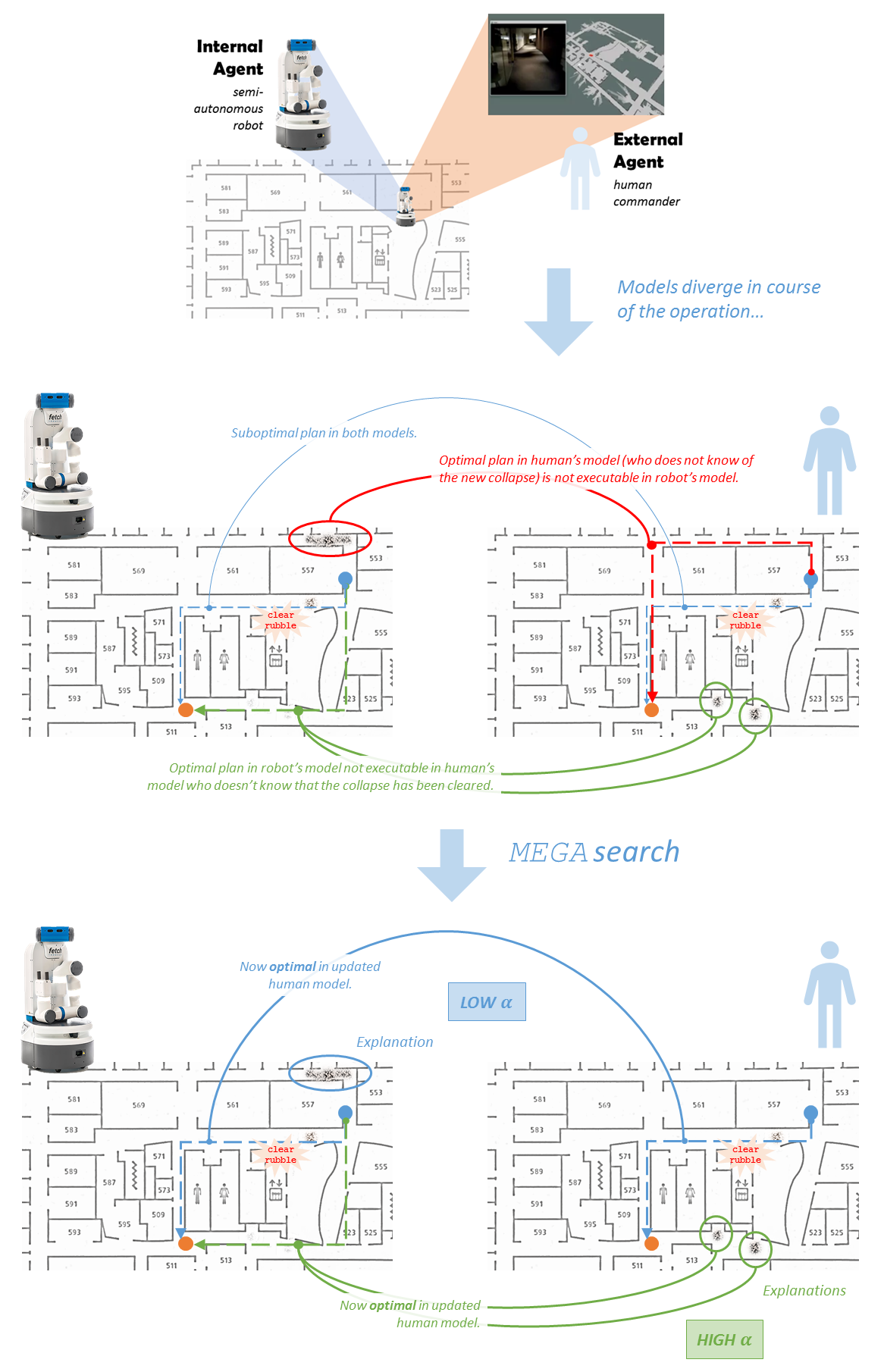}
\caption{A typical search and reconnaissance scenario with an internal semi-autonomous agent (robot) and an external supervisor (human) -- 
a video demonstration can be accessed at \textcolor{blue}{\url{https://www.youtube.com/watch?v=u_t1TQotzo4}}.
}
\label{map}
\end{figure}

\subsection{Demonstration: The USAR Domain}

We first demonstrate \mega~on a robot performing an Urban Search And Reconnaissance (USAR) task -- here a remote robot is put into disaster response operation often controlled partly or fully by an external human commander. This is a typical USAR setup \cite{nancy}, where the robot's job is to infiltrate areas that may be otherwise harmful to humans, and report on its surroundings as and when required / instructed by the external. The external usually has a map of the environment, but this map is no longer accurate in a disaster setting - e.g. new paths may have opened up, or older paths may no longer be available, due to rubble from collapsed structures like walls and doors. The robot (internal) however may not need to inform the external of all these changes so as not to cause information overload of the commander who may be otherwise engaged in orchestrating the entire operation. This calls for an instantiation of the \mega~algorithm where the model differences are contributed to by changes in the map, i.e. the initial state of the planning problem (the human model has the original unaffected model of the world). 

Figure \ref{map} shows a relevant section of the map of the environment where this whole scenario plays out. The orange marks indicate rubble that has blocked a passage, while the green marks indicate collapsed walls.
The robot (Fetch), currently located at the position marked with a blue {\bf \textcolor{blue}{O}}, is tasked with taking a picture at location marked with an orange {\bf \textcolor{orange}{O}} in the figure. 
The external commander's expects the robot to take the path shown in red, which is no longer possible. The robot armed with \mega~ has two choices -- it can either follow the green path and explain the revealed passageway due to the collapse, or compromise on its optimal path, clear the rubble and proceed along the blue path. 
{\em A video demonstration of the scenario can be viewed at \textcolor{blue}{\url{https://www.youtube.com/watch?v=u_t1TQotzo4}}.}
The first part of the video demonstrates the plan generated by \mega~for low $\alpha$ values. As expected, it chooses the blue path that requires the least amount of explanation, and is thus the most explicable plan. In fact, the robot only needs to explain a single initial state change to make this plan optimal, namely --

\vspace{5pt}
{
\begin{verbatim}
Explanation >> remove-has-initial-state-clear_path p1 p8
\end{verbatim}
}

This is also an instance where the plan closest to the human expectation, i.e. the most explicable plan, still requires an explanation, which previous approaches in the literature cannot provide. Moreover, in order to follow this plan, the robot must perform the costly \texttt{clear\_passage p2 p3} action to traverse the corridor between \texttt{p2} and \texttt{p3}, which it could have avoided in its optimal plan (shown in green on the map). Indeed, \mega~switches to the robot's optimal plan for higher values of $\alpha$ along with the following explanation --

\vspace{5pt}
{
\begin{verbatim}
Explanation >> add-has-initial-state-clear_path p6 p7
Explanation >> add-has-initial-state-clear_path p7 p5
Explanation >> remove-has-initial-state-clear_path p1 p8
\end{verbatim}
}

By providing this explanation, the robot is able to convey to the human the optimality of the current plan as well as the infeasibility of the human's expected plan (shown in red).

\subsection{Human Factors Evaluations}

Finally, we will now use the above search and reconnaissance domain to analyze how humans respond to the explicability versus explanations trade-off. This is done by exposing the external commander's interface to participants who get to analyze plans in a mock USAR scenario. 
The participants were incentivized to make sure that the explanation does indeed help them understand the optimality of the plans in question by formulating the interaction in the form of a game. 
This is to make sure that participants were sufficiently invested in the outcome as well as mimic the high-stakes nature of USAR settings to accurately evaluate the explanations.

\begin{figure}
\includegraphics[width=\columnwidth]{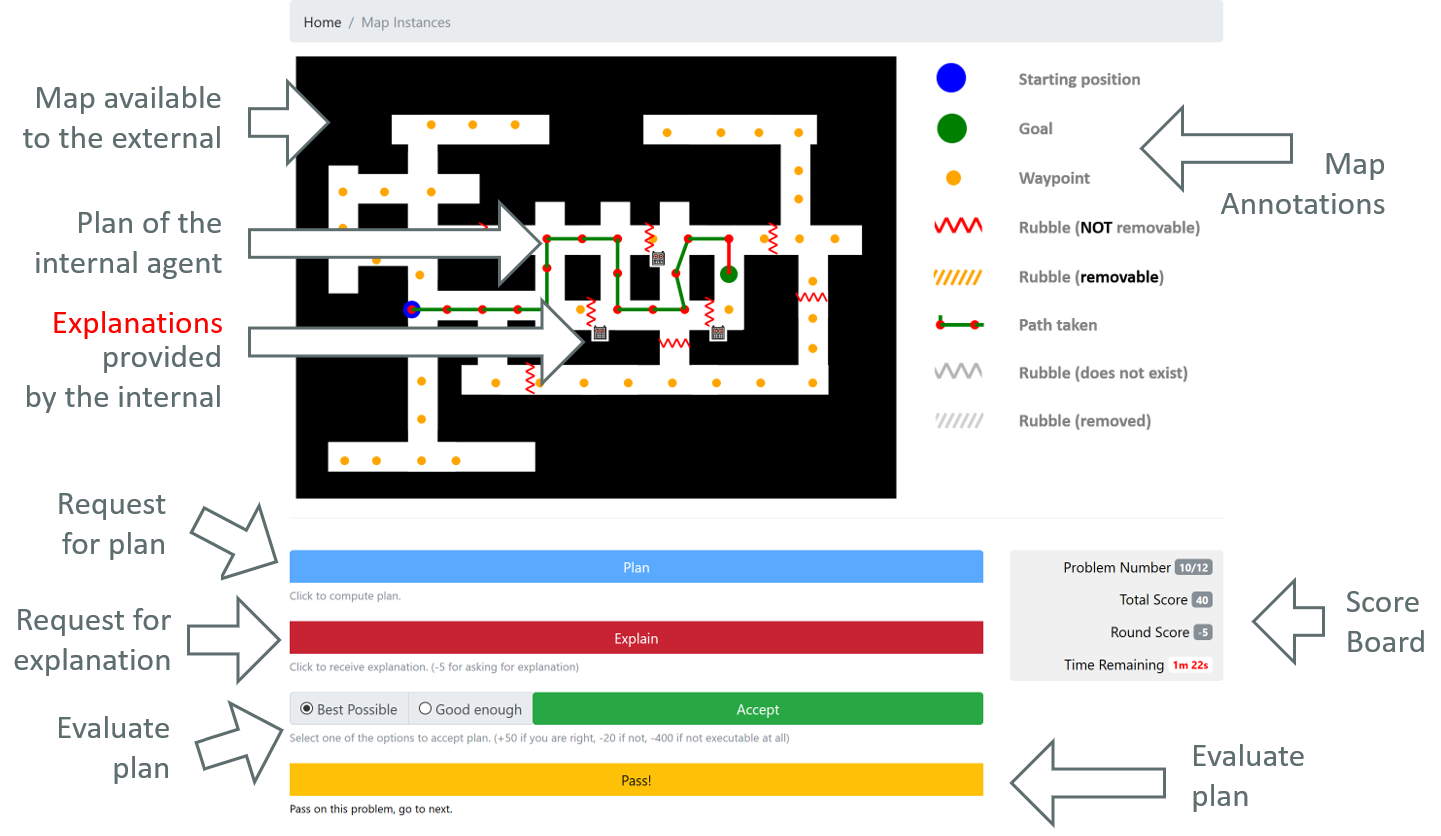}
\caption{Interface to the external commander in a mock search and reconnaissance study.
}
\label{kitty}
\end{figure}

Figure~\ref{kitty} shows a screenshot of the interface which displays to each participant an initial map (which they are told may differ from the robot's actual map), the starting point and the goal. 
A plan is illustrated in the form of a series of paths through various waypoints highlighted on the map. 
The participant has to identify if the plan shown is optimal. 
If the player is unsure, they can ask for an explanation. 
The explanation is provided to the participant in the form of a set of model changes in the player's map. 

The scoring scheme for the game is as follows. 
Each player is awarded 50 points for correctly identifying the plan as either optimal or satisficing. 
Incorrectly identification costs them 20 points. 
Every request for explanation further costs them 5 points, while skipping a map does not result in any penalty. 
The participants were additionally told that selecting an inexecutable plan as either feasible or optimal would result in a penalty of 400 points. 
Even though there were no actual incorrect plans in the dataset, this information was provided to deter participants from taking chances with plans they did not understand well.

Each participant was paid \$10 dollars and received additional bonuses based on the following payment scheme --

\begin{itemize}
\item[-] Scores higher than or equal to 540 were paid \$10.
\item[-] Scores higher than 540 and 440 were paid \$7.
\item[-] Scores higher than 440 and 340 were paid \$5.
\item[-] Scores higher than 340 and 240 were paid \$3.
\item[-] Scores below 240 received no bonuses.
\end{itemize}

The scoring systems for the game was designed to make sure -- 

\begin{itemize}
\item Participants should only ask for an explanation when they are unsure about the quality of the plan (due to small negative points on explanations).\\[-2ex]
\item Participants are incentivized to identify the feasibility and optimality of the given plan correctly (large reward and penalty on doing this wrongly).
\end{itemize}

Each participant was shown a total of 12 maps. 
For 6 of the 12 maps, the participant was assigned the optimal robot plan, and when they asked for an explanation, they were randomly shown different types of explanations as introduced in \cite{explain}. 
For the rest of the maps, in place of the robot's optimal plan, participants could potentially be assigned a plan that is optimal in the human model (i.e. an explicable plan) with no explanation or somewhere in between (i.e. the balanced plan) with a shorter explanation.
Note that out of the 6 maps, only 3 had both balanced plans as well as explicable plans, the other 3 either had a balanced plan or the optimal human plan. 
In total, we had 27 participants for the study, including 4 female and 22 male participants between the age range of 19-31 (1 participant did not reveal their demographic).

\begin{figure}[tbp!]
\includegraphics[width=\columnwidth]{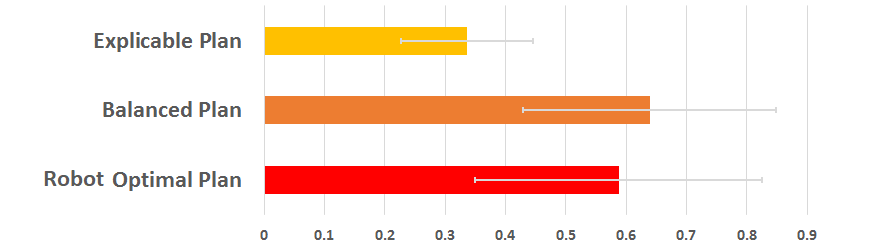}
\caption{Responses to explicable plans versus balanced or robot optimal plans with explanations.}
\label{fig1}
\end{figure}

\begin{figure}[tbp!]
\includegraphics[width=\columnwidth]{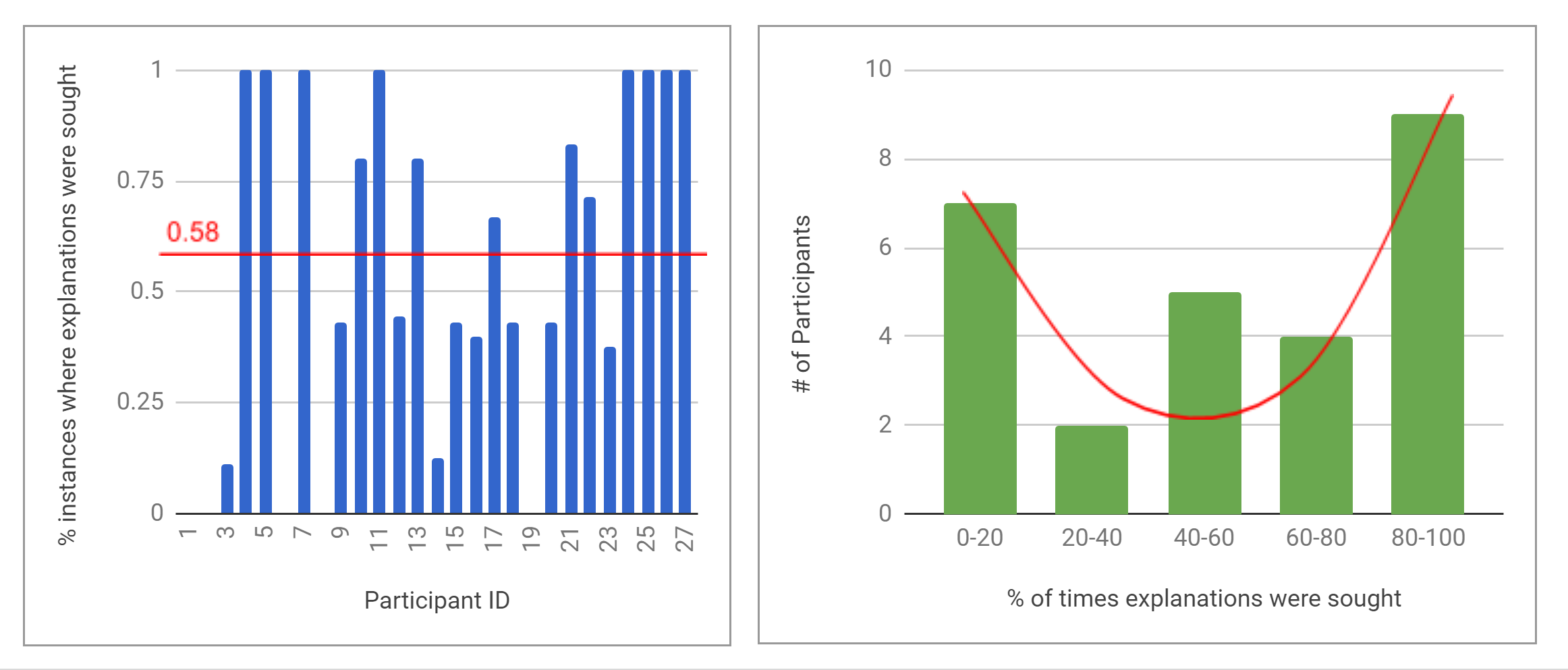}
\caption{Click-through rates for explanations.}
\label{fig2}
\end{figure}

\begin{table}[tbp!]
\centering\small
\begin{tabular}{|c|c||c|c||c|c|}
\hline
\multicolumn{2}{c}{Optimal Plan} & \multicolumn{2}{c}{Balanced Plan} & \multicolumn{2}{c}{Explicable Plan} \\
\hline
$|\mathcal{E}|$ & $C(\pi, \mathcal{M}^R)$ & $|\mathcal{E}|$ & $C(\pi, \mathcal{M}^R)$ & $|\mathcal{E}|$ & $C(\pi, \mathcal{M}^R)$ \\
\hline
2.5 & 5.5 & 1 & 8.5 & - & 16\\
\hline
\end{tabular}
\vspace{10pt}
\caption{Statistics of explicability versus explanation trade-off with respect to explanation length and plan cost.}
\label{tab1}
\end{table}

Figure~\ref{fig1} shows how people responded to the different kinds of explanations / plans. These results are from 382 problem instances that required explanations, and 25 and 40 instances that contained balanced and explicable plans respectively. From the perspective of the human, the balanced plan and the robot optimal plan do not make any difference since both of them appear suboptimal. This is evident from the fact that the click-through rate for explanations in these two conditions are similar. However, the rate of explanations is significantly less in case of explicable plans as desired. 

Table~\ref{tab1} shows the statistics of the explanations / plans. These results are from  124 problem instances that required {\em minimal} explanations as per \cite{explain}, and 25 and 40 instances that contained balanced and explicable plans respectively, as before.
As desired, the robot gains in length of explanations but loses out in cost of plans produced as it progresses along the spectrum of optimal to explicable plans. Thus, while Table~\ref{tab1} demonstrates the cost of explanation versus explicability trade-off from the robot's point of view, Figure~\ref{fig1} shows how this trade-off is perceived from the human's perspective.

It is interesting to see that in Figure~\ref{fig1} about a third of the time participants still asked for explanations even when the plan was explicable, and thus optimal in their map. This is an artifact of the risk-averse behavior incentivized by the gamification of the explanation process and indicative of the cognitive burden on the humans who are not (cost) optimal planners. Thus, going forward, the objective function should incorporate the cost or difficulty of analyzing the plans and explanations from the point of view of the human in addition to the current costs in equation \mega(4) and Table~\ref{tab1} modeled from the perspective of the robot model.

Finally, in Figure~\ref{fig2}, we show how the participants responded to inexplicable plans, in terms of their click-through rate on the explanation request button. Such information can be used to model the $\alpha$ parameter to situate the explicability versus explanation trade-off according to preferences of individual users. 
It is interesting to see that the distribution of participants (right inset) seem to be bimodal indicating that there are people who are particularly skewed towards risk-averse behavior and others who are not, rather than a normal distribution of response to the explanation-explicability trade-off. This further motivates the need for learning $\alpha$ interactively with the particular human in the loop.

\section{Discussion and Future Work}

In the following section, we will elaborate on some of the exciting avenues of future research borne out of this work.

\subsection{Model learning and picking the right $\alpha$}

We assumed that the hyper-parameter $\alpha$ is set by the designer in determining how much to trade-off the costs of explicability versus explanations on the part of the autonomous agent. However, the design of $\alpha$ itself can be more adaptive and ``human-aware'' in the sense that the parameter can be \emph{learned} in course of interactions with the human in the loop to determine what kind of plans are preferred (as seen in Figure~\ref{fig2}) and how much information can be transmitted. 
This is also relevant in cases where the human mental model is not known precisely or if there is uncertainty towards what the new model is after an update or explanation. 
This is a topic of future work; existing literature on iterative model learning \cite{nikolaidis2015improved,hadfield2016cooperative} can provide useful guidance towards the same.
Authors in \cite{chakraborti2017ai} discuss a few useful representations for learning such models for the purposes of task planning at various levels of granularity. 
Note that search with uncertainty over a learned human (mental) model can often times be compiled to the same planning process as described in \cite{exp-fss-multi} by using annotated models, so the same techniques as introduced in this paper still apply.

\subsection{Cost of explanations and cognitive load}

Currently, we only considered the cost of explanations and explicability from the point of view of the robot. 
However, there might be additional (cognitive) burden on the human -- measured in terms of the complexity of interpreting an explanation and how far away the final plan is from the optimal plan in the human's mental model. 
This again ties back to the assumptions on the cognitive abilities (i.e. optimality) of the human in the loop, and needs calibration \cite{nikolaidis2015improved,hadfield2016cooperative} based on repeated interactions (as seen in Figure~\ref{fig1}). 

\section{Conclusion}

We saw how an agent can achieve human-aware behavior while at the same time keeping in mind the cost of departure from its own optimality which could otherwise have been explained away if given the opportunity. This raises several intriguing challenges in the plan generation process, most notably in finding better heuristics in speeding up the model space search process as well as dealing with model uncertainty and identifying the sweet spot of the algorithm in explicability-explanations trade-off. Indeed, the revised human-aware planning paradigm opens up exciting new avenues of research such as learning human mental models, providing explanations at different levels of abstractions, and so on.


\bibliographystyle{ACM-Reference-Format}  
\bibliography{bib}  

\end{document}